\pdfoutput=1

\documentclass[11pt]{article}

\usepackage[]{acl}

\usepackage{times}
\usepackage{latexsym}

\usepackage[T1]{fontenc}

\usepackage[utf8]{inputenc}

\usepackage{microtype}
\usepackage{booktabs}
\usepackage{amsmath,amssymb,amsfonts}
\usepackage{pifont}
\usepackage{bbding}
\usepackage{multirow}
\usepackage{multicol}
\usepackage{diagbox}
\usepackage{makecell}
\usepackage{graphicx}

\title{Ask Question First for Enhancing Lifelong Language Learning}

\author{Han Wang\textsuperscript{1, 2} \and 
Ruiliu Fu\textsuperscript{1, 2}\and 
Xuejun Zhang\textsuperscript{1, 2}\and
Jun Zhou\textsuperscript{1, 2}\and
Qingwei Zhao\textsuperscript{1, 2}\\
\textsuperscript{1}Institute of Acoustics, Chinese Academy of Sciences, Beijing, China. \\
\textsuperscript{2}University of Chinese Academy of Sciences, Beijing, China\\
\texttt{\{wanghan, furuiliu, zhangxuejun, zhoujun, qzhao\}@hccl.ioa.ac.cn} \\}

\begin{document}
\maketitle
\begin{abstract}
Lifelong language learning aims to stream learning NLP tasks while retaining knowledge of previous tasks. Previous works based on the language model and following data-free constraint approaches have explored formatting all data as "begin token (\textit{B}) + context (\textit{C}) + question (\textit{Q}) + answer (\textit{A})" for different tasks. However, they still suffer from catastrophic forgetting and are exacerbated when the previous task's pseudo data is insufficient for the following reasons: (1) The model has difficulty generating task-corresponding pseudo data, and (2) \textit{A} is prone to error when \textit{A} and \textit{C} are separated by \textit{Q} because the information of the \textit{C} is diminished before generating \textit{A}. Therefore, we propose the Ask Question First and Replay Question (AQF-RQ), including a novel data format "\textit{BQCA}" and a new training task to train pseudo questions of previous tasks. Experimental results demonstrate that AQF-RQ makes it easier for the model to generate more pseudo data that match corresponding tasks, and is more robust to both sufficient and insufficient pseudo-data when the task boundary is both clear and unclear. AQF-RQ can achieve only 0.36\% lower performance than multi-task learning.

\end{abstract}

\section{Introduction}


Lifelong learning is the capacity of human beings to acquire, reconstruct, strengthen, and transfer knowledge \citep{Ring97}. Human beings can learn new knowledge while consolidating old knowledge by first detecting and learning the distinctions between old and new knowledge and then simplifying the old and new knowledge based on leveraging the common points. This concept is critical for encouraging machines to learn NLP tasks in a similar way that people do. In the application of NLP, new data are continuously acquired and categorized as either new data for existing tasks or new data for new tasks. 
\begin{figure}[t]
\centering 
\includegraphics[scale=0.4]{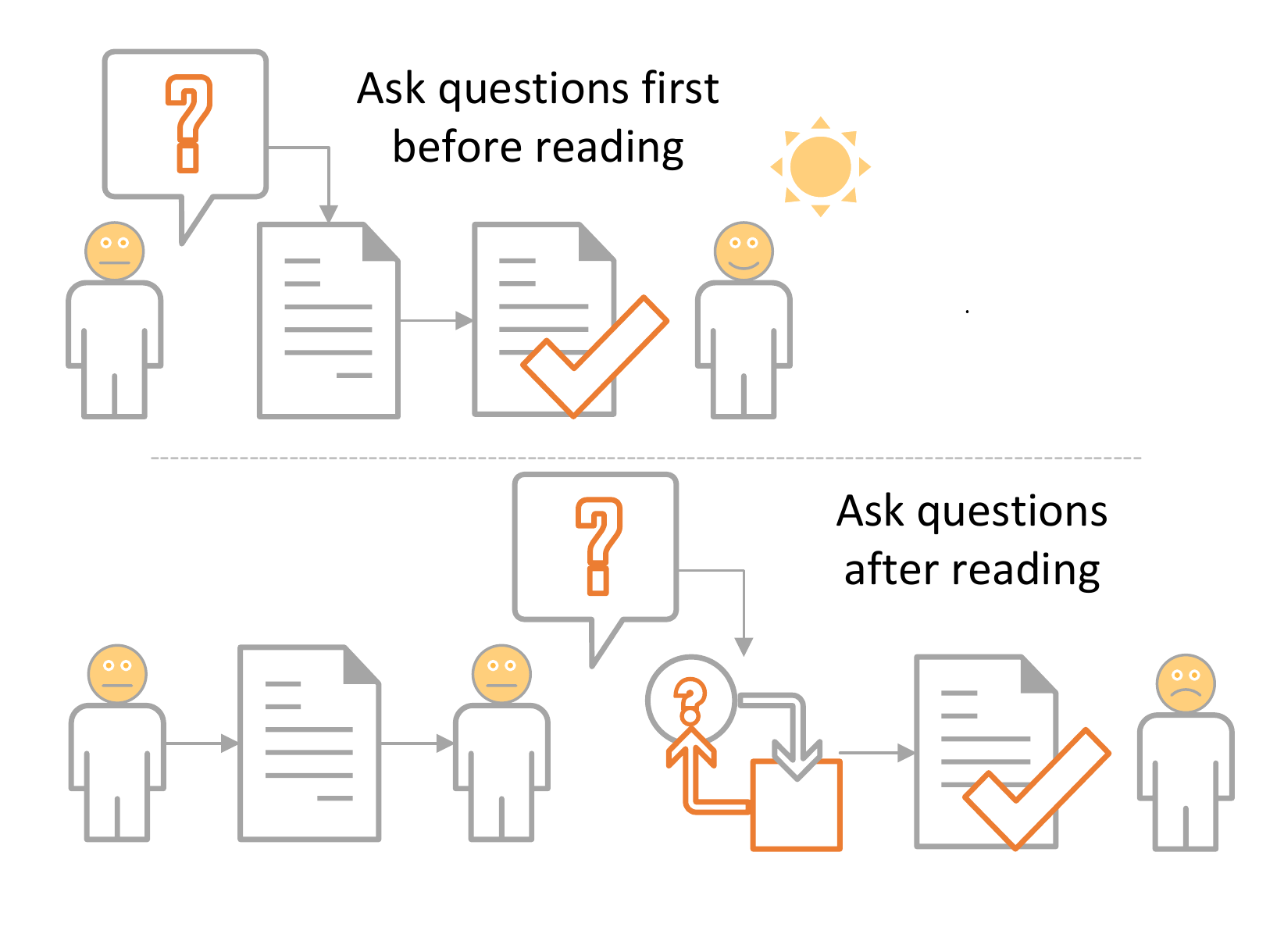} 
\caption{Illustration of the difference between asking questions before reading and asking after reading.}
\label{fig:aqf}
\end{figure}For new data of existing tasks, the traditional method, known as isolated learning \citep{lll-book-2nd}, is to retrain the model with the old data appended with new data. For data from new tasks, multi-task learning (MTL) integrates data from previous and new tasks to retrain the model. Both of these methods are limited to assuming that all new and old data can be obtained during training. However, the reality is that tasks are acquired and trained in the stream, which makes the model suffer from catastrophic forgetting \citep{Ring97,mccloskey1989catastrophic,french1999catastrophic} (i.e. forgetting previously learned tasks/knowledge).

Lifelong language learning (LLL), which this paper focuses on, aims to learn a stream of NLP tasks with lifelong learning. LAMOL \citep{sun2019lamol} has recently proposed implementing a language model for LLL by formatting all data as QA-style and generating pseudo data instead of real data for previous tasks in order to prevent catastrophic forgetting. Many works \citep{l2kd,dnr,rational-lamol} have investigated how to improve it through methods that require more computing resources or additional parameters (e.g. knowledge distillation or adding sub-networks). In this paper, we refer to these works as LAMOL-based methods. In LAMOL-based methods, each example is consist of four segments: task-specific or task-independent token [TASK]/[GEN] (\textit{B}), context (\textit{C}), question (\textit{Q}), and answer (\textit{A}). Then each example is formatted as "\textit{B+C+Q+A}" ("\textit{BCQA}"). LAMOL has a gap between MTL that is regarded as the upper bound of LLL. Other LAMOL-based methods require more computing resources or additional parameters but still have a gap between MTL especially pseudo samples is unsufficient. Therefore, the "\textit{BCQA}" format is useful but not train-efficient enough for reasons below: (1) generating task-corresponding pesudo data is hard because the similar \textit{C} can be found in distinct tasks; (2) the \textit{C}'s information is diminished when \textit{A} is generated due to the fact that \textit{A} and \textit{C} are separated by \textit{Q}.

To generate more task-corresponding pseudo data for previous tasks and tighten the relationship between the \textit{C} and the \textit{A}, we proposed the Ask Question First (AQF) and Replay Question (RQ).

The AQF formats all data into a novel format: "\textit{B+Q+C+A}" ("\textit{BQCA}"), which is consistent with human reading comprehension behavior. Questions are more than just questions; they direct our learning. As is shown in Fig.\ref{fig:aqf}, when people do reading comprehension, they usually read the questions first, then read the articles with the questions, and pay attention to what can answer the questions while reading. This is an efficient reading and learning method, which known as metacognition \citep{Flavell1979MetacognitionAC}, has been researched in the field of education and psychology. There are two benefits to applying the "\textit{BQCA}" format: (1) It's easier to generate pseudo-data that matches the corresponding task because the \textit{Q} has different but limited types for each task. (2) The model is more stable when there isn't enough pseudo data because the \textit{A} is right next to the \textit{C}. Furthermore, the model can pay more attention to important information in the \textit{C} with the help of the \textit{Q}.

The RQ introduce a novel training task to help model generate more task-corresponding pseudo data. In the "\textit{BQCA}" format, generating the correct \textit{Q} is crucial for generating task-corresponding pseudo-data. However, the \textit{Q} of the previous may be covered by the new task since we cannot predict the number and type of \textit{Q} of the new task. In order to strengthen the generation of the \textit{Q} of the previous task, we generate pseudo-problems of the previous task to train the model.

The contributions of our paper are listed below: 
(1) We proposed the Ask Question First and Replay Question (AQF-RQ\footnote{\url{https://github.com/CodeHan/AQF-RQ}}) to alleviate catastrophic forgetting when the pseudo data is sufficient and insufficient without additional computation resources and parameters. 
(2) We proposed a novel data format "\textit{BQCA}" to make data train-efficient and generating corresponding pseudo data easier. 
(3) We proposed a novel training task to help the model generate correct questions and then generate more task-corresponding pseudo data.

\section{Related Work}
Lifelong language learning (LLL) is an essential step in promoting the realization of general artificial intelligence in the field of NLP. \citep{lll-2019NAACL-SR,lll-emmlp2020-dialogue,lll-naacl2021-tc} have studied LLL on a single type of NLP task by regularization or replaying real data.  Recently, LAMOL \citep{sun2019lamol} uses a language model (LM) to learn various kinds of NLP tasks in QA-style. In LAMOL, the pseudo-data generated by the model is trained together with the new task to alleviate catastrophic forgetting. Many works explored the enhancement of LAMOL with additional computation resources or adding sub-networks. L2KD \citep{l2kd}, DnR \citep{dnr} and DFSD \citep{lll-DFSD} distilled parts or all layers of the model to improve LAMOL. ARPER \citep{lll-emmlp2020-dialogue} applied regualarization on parameters with prioritized exemplar replay. Rational-LAMOL \citep{rational-lamol} applied critical freezing guided by rationale information which is obtained by human or unsupervised rationale generation \citep{2020ICML-unsupervise-rationale}. \citep{lll-emnlp-adapter} applied Adapter \citep{lll-icml2019-adapter} to plug into pretrain language model. Those are effective but require more computational resources (e.g., distillation and regularization) or new parameters (such as adding sub-networks for new tasks). We aspire to improve without increasing our resources. Consequently, based on the metacognition \citep{Flavell1979MetacognitionAC, metacog-2010, metacog-cogsci2021} proposed in education and psychology, we propose a more training-efficient and robust data format: first ask questions, then observe the context, finally answer the questions. In addition, we introduce replay questions to strengthen the model's attention to questions.

\section{Methodology}
In this section, we first introduce LAMOL in Section \ref{sec:lamol}. Then, our proposed Ask Question First (AQF) is detailed in Section \ref{sec:aqf}. Thirdly, we introduce the Replay Questions in Section \ref{sec:replay question}. Finally, we summarize the training objectives in our paper in Section \ref{sec:loss}.

\begin{figure}[t]
\centering 
\includegraphics[scale=0.3]{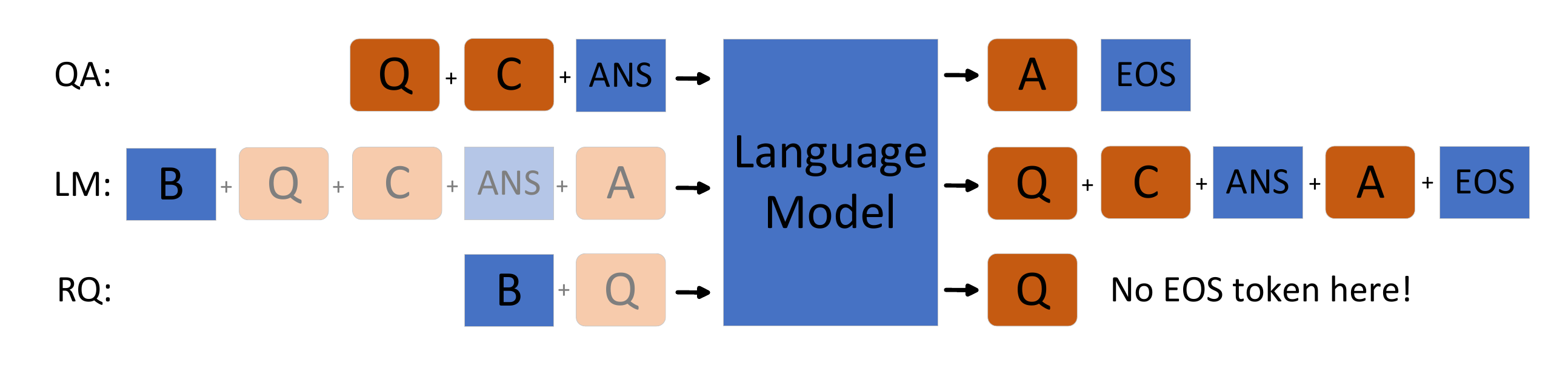} 
\caption{Illustration of QA/LM/RQ task with the "\textit{BQCA}" format.}
\label{fig:all tasks}
\end{figure}

\subsection{LAMOL}\label{sec:lamol}
LAMOL \citep{sun2019lamol} proposed using a single language model for lifelong language learning. A stream of various NLP tasks is learned by GPT-2 \cite{gpt2} with joint training of the language model (LM) task and the question-answering (QA) task. In LAMOL, all tasks are formatted QA-style. Each example can consist of four segments: task-specific or task-independent token [TASK]/[GEN] (\textit{B}), context (\textit{C}), question (\textit{Q}), and answer (\textit{A}). In this way, the LM task is to generate "\textit{C+Q+A}" by inputting [TASK]/[GEN], and the QA task is to generate "\textit{A}" by inputting "\textit{C+Q}". The loss of LAMOL is calculated as below:
\begin{equation}
\mathcal{L}_{LAMOL} = \mathcal{L}_{QA}+\lambda \mathcal{L}_{LM}
\end{equation}
where $\lambda$ is the weight of the LM task.

With the help of the LM task, the [TASK]/[GEN] can be used as the first token to input the model and generate the pseudo sample by greedy decoding. From the second task on, pseudo samples of previous tasks are generated in this way and jointly trained with the new task dataset to alleviate catastrophic forgetting. Let denote the set of pseudo samples as $\mathcal{P}_i = \{ ps_{j}|i,j \in N_+, 1 \leq j<i \}$, where $ps_{j}$ represents the pseudo data of $j$-th task.

The number of pseudo samples of previous tasks is determined by the size of the new task and a hyperparameter sampling ratio $\gamma \in (0,1]$. Assuming that $T = \{T_{1},T_{2},…,T_{N}\}$ represents N tasks to be learned. Let $|D_{t}|, t \in [2, N]$ denote the size of $t$-th task. LAMOL generates $|ps_i|= \frac{\gamma}{t-1}|D_{t}|$ pseudo-samples for each previously learned task. Then, $D_t$ is joined with $\mathcal{P}_t$ to train the model. However, the performance of LAMOL still has a gap between multi-task learning (MTL), which is regarded as the upper bound of lifelong learning, since not each pseudo sample is tied to [TASK]/[GEN]. In this paper, we propose Ask Question First (AQF) and replaying questions to make data train-efficient and shorten the gap between MTL.

\begin{table}[t]
  \centering
    \begin{tabular}{c|c|c}
    \toprule
    Case ID & Content & Cor.\\
    \midrule
    1     & \textit{B}$^{T_1}+$\textit{C}$^{T_1}+$\textit{Q}$^{T_1}+$\textit{A}$^{T_1}$ &\ding{51}\\
    2     & \textit{B}$^{T_1}+$\textit{C}$^{T_2}+$\textit{Q}$^{T_2}+$\textit{A}$^{T_2}$ &\ding{55}\\
    3     & \textit{B}$^{T_2}+$\textit{C}$^{T_2}+$\textit{Q}$^{T_2}+$\textit{A}$^{T_2}$ &\ding{51}\\
    4     & \textit{B}$^{T_2}+$\textit{C}$^{T_1}+$\textit{Q}$^{T_1}+$\textit{A}$^{T_1}$ &\ding{55}\\
    \bottomrule
    \end{tabular}%
    \caption{The cases of the pseudo sample after the model learning two tasks in turn with "\textit{BCQA}" format. Cor. is the abbreviation of "corresponding".}
  \label{tab:pseudo-case}%
\end{table}%

\subsection{Ask Question First}\label{sec:aqf}
Ask Question First (AQF) is a novel data format that makes the model imitate the process of human reading comprehension to learn knowledge from the training set efficiently. LAMOL proposed a data format, "\textit{B+C+Q+A}" ("\textit{BCQA}"), which is useful but not train-efficient enough. 
All pseudo samples are generated starting with [TASK]/[GEN]. However, [TASK]/[GEN] is not strongly tied with the \textit{C} of the tasks because the context is complex and volatile for the task. Different tasks can have similar contexts. Therefore, [TASK]/[GEN] is easy to be biased to the new task resulting in pseudo samples not corresponding to their tasks when more tasks are learned. These non-corresponding pseudo-samples jointly trained with the new task will aggravate the catastrophic forgetting of the model. Otherwise, the \textit{C}'s information is diminished when \textit{A} is generated due to the fact that \textit{A} and \textit{C} are separated by \textit{Q}.

To generate more corresponding pseudo samples, we propose AQF format all data as "\textit{B+Q+C+A}" ("\textit{BQCA}") which is simple yet efficient to make the model learn knowledge from the training set efficiently. We choose the task-specific token to analyze why our proposed "\textit{BQCA}" is better than the naive "\textit{BCQA}". Assuming that the model has learned two tasks $T_1$ and $T_2$, pseudo samples of $T_1$ and $T_2$ needed to be generated before training the new task. Two types of task-specific tokens \textit{B}$_1$ and \textit{B}$_2$ are inputted to the model to generate pseudo samples with greedy decoding. There are four main kinds of pseudo samples that can be obtained as shown in Table \ref{tab:pseudo-case}. Case 1 and Case 3 are nice pseudo samples that can alleviate catastrophic forgetting, but Case 2 and Case 4 are terrible ones that can aggravate catastrophic forgetting. Ideally, pseudo samples, which we denote as task-corresponding pseudo-samples $\mathcal{P}_c$, like Case 1 and Case 3 are what we expect to generate. However, in fact, due to catastrophic forgetting, Case 2 and Case 4 also appear. These pseudo samples like case 2 and case 4 are denoted as the not task-corresponding samples $\mathcal{P}_{nc}$. The probability of generating $\mathcal{P}_{nc}$ is positively correlated with the severity of catastrophic forgetting. Therefore, the model will form a vicious circle: forgetting old tasks and generating $\mathcal{P}_{nc}$ with greater probability. For pseudo samples start with task-specific token, the $ps_i$ consists of corresponding pseudo data $ps_i^c$ and not corresponding pseudo data $ps_i^{nc}$ (i.e. $ps_i=\{ps_i^c,ps_i^{nc}\}$).

In LAMOL-based methods, they used GPT-2 \citep{gpt2} for experiments. GPT-2 is an autoregressive model that can only obtain information before the current position, and then predict the words. Therefore, the probability of our proposed "\textit{BQCA}" format and naive "\textit{BCQA}" format can be calculated by Eq.(\ref{eq:qca}) and Eq.(\ref{eq:cqa}), respectively:
\begin{align}
        P_{AQF}&=P(BQCA) \label{eq:qca}\\
         &= P(A|BQC)P(C|BQ)P(Q|B)P(B) \nonumber\\
		P_{naive}&=P(BCQA) \label{eq:cqa}\\
         &= P(A|BCQ)P(Q|BC)P(C|B)P(B)\nonumber
\end{align}



As shown in Eq.(\ref{eq:qca}) and Eq.(\ref{eq:cqa}), $P(B)$ is a constant because it is given as the first token. Next, $P(Q|B)$ and $P(C|B)$ affect the quality of the generated pseudo data. To obtain $\mathcal{P}_c$, the \textit{Q} or \textit{C} should be generated to be corresponding to the given \textit{B}, then the remaining generation has a greater probability correspond to \textit{B}. 

Then we first consider $P(Q|B)$ and $P(C|B)$. Let $\Theta_N$ denote the parameters of the model after learning N tasks in turn. Assuming two tasks have been learned on the condition of sufficient pseudo samples (i.e. large $\gamma$) in turn, $P(Q|B)$ and $P(C|B)$ can be written as, $P(Q|B,\Theta_2)$ and $P(C|B,\Theta_2)$. In the dataset of a task, there are thousands of kinds of \textit{C}, but there are only limited kinds of \textit{Q}. For a simplest classification task, only one fixed \textit{Q} is needed. For example, in the SST \citep{radford2017sst}, all \textit{Q} is "\textit{is review is positive or negative?}". For complex tasks, such as SRL \citep{he2017srl}, it is mainly based on limited types of \textit{Q} such as \textit{what}, \textit{where}, \textit{why}, \textit{when}, \textit{how} etc, but it is still smaller than that of \textit{C}. Therefore, $P(Q|B,\Theta_2)$ and $P(C|B,\Theta_2)$ are inversely proportional to the vocabulary sizes of \textit{Q} and \textit{C}, respectively. Let $\mathcal{V}_Q=\{w_i^Q|i \in [1,|\mathcal{V}_Q|]\}$ and $\mathcal{V}_C=\{w_i^C|i \in [1,|\mathcal{V}_C|]\}$ denote the vocabulary of \textit{Q} and \textit{C}, respectively. $w_i^Q$ and $w_i^C$ denotes the word in the vocabulary of \textit{Q} and \textit{C}, respectively. According to the above analysis, it can be obtained that $|\mathcal{V}_Q| \ll |\mathcal{V}_C$|. For most words, since their probability is inversely proportional to the size of the vocabulary, we can conclude that $P(w_Q) \gg P(w_C)$. Slight noise perturbations, such as shifts from the data distribution of the new task, can make $P(w_C)$ more susceptible than $P(w_Q)$. On the other hand, different tasks may have similar \textit{C} due to a large number of similar common phrases, but the \textit{Q} is almost different in what the questions pay attention to. 
Therefore, we can conclude that:
\begin{align}
     P_{B_1}(w_Q^{T_1}) \geq P_{B_1}(w_Q^{T_2})>P_{B_1}(w_C^{T_1}) \geq P_{B_1}(w_C^{T_2}) \label{eq:b1}\\
    P_{B_2}(w_Q^{T_2}) \geq P_{B_2}(w_Q^{T_1})>P_{B_2}(w_C^{T_2}) \geq P_{B_2}(w_C^{T_1})
\end{align} 
where the subscript $B_1$ or $B_2$ means the probability on the condition of $B_1$ or $B_2$. Then, we can conclude that:
\begin{align}
P(Q^{T_1}|B^{T_1},\Theta_2) > P(C^{T_1}|B^{T_1},\Theta_2) \label{eq:q1_c1}\\
P(Q^{T_2}|B^{T_2},\Theta_2) > P(C^{T_2}|B^{T_2},\Theta_2) \label{eq:q2_c2}
\end{align}
It can be concluded that generating \textit{Q} first is more stable than generating \textit{C}.

On the condition that the \textit{Q} or \textit{C} is corresponding to the B, we analysis the $P(C|BQ,\Theta_2)$ and $P(Q|BC,\Theta_2)$. With the help of GPT-2, a Transformer decoder-based model, the current token can pay more attention to the information before the current position. When the previous adjacent content contains enough information, the generated token strongly correlates with the previous content. Therefore, $P(C^{T_1}|B^{T_1}Q^{T_1},\Theta_2)$ is close to $P(Q^{T_1}|B^{T_1}C^{T_1},\Theta_2)$, and $P(C^{T_2}|B^{T_2}Q^{T_2},\Theta_2)$ is close to $P(Q^{T_2}|B^{T_2}C^{T_2},\Theta_2)$.

Finally, $P(A|BQC)$ and the $P(A|BCQ)$ are trained with the QA task. It can be ignored from the perspective of generating task-corresponding pseudo-samples. However, correct "\textit{A}" is easier to obtain on the condition of correct "\textit{BQC}" than correct "\textit{BCQ}" because the "\textit{C}" is right next to "\textit{A}" in "\textit{BQC}". In "\textit{BQC}", "\textit{Q}" can help the model pay more attention to the important information in "\textit{C}" and then have a larger probability of generating the correct "\textit{A}".

From the above analysis, we can conclude that $P(Q|B)$ and $P(C|B)$ play main role on generating task-corresponding pseudo-samples $\mathcal{P}_c$ according to $P_{AQF}$ and $P_{naive}$, respectively. As shown in Eq.(\ref{eq:q1_c1}) and Eq.(\ref{eq:q2_c2}), $P(Q|B)$ are better than $P(C|B)$ for generating $\mathcal{P}_c$ after sufficient pseudo samples being joint trained.

\subsection{Replay Questions}\label{sec:replay question}
Replay Questions (RQ) is proposed to strengthen the probability of generating task-corresponding \textit{Q} when given \textit{B}. Based on Eq.(\ref{eq:b1}), $P_{B_1}(w_Q^{T_1})$ greater than or equal to $P_{B_1}(w_Q^{T_2})$ on the condition of sufficient pseudo samples being joint trained. However, the situation $P_{B_1}(w_Q^{T_2})>P_{B_1}(w_Q^{T_1})$ may happen when the amount of pseudo samples is much less than that of the new task (i.e. $|D_{2}|\gg \gamma|D_{2}|$). Since $P(Q|B)$ plays a main role in $P_{AQF}$, we propose to replay questions to make $P(Q_i|B_i)$ larger than $P(Q_{j\neq i}|B_i)$.

As shown in Fig.\ref{fig:all tasks}, RQ is to generate the questions of pseudo samples that start with [TASK]/[GEN] but have no end token. Since the object of RQ is to make the model generate task-corresponding pseudo-samples, RQ should not supply the information about the end of the sentence.

\subsection{Training}\label{sec:loss}
In summary, our proposed AQF is applied to both the LM task and the QA task. Otherwise, we expand the training objective with our proposed RQ. The final training objective is shown as Eq.(\ref{eq:final_loss}).
\begin{equation}
    \mathcal{L} = \mathcal{L}_{QA}^{AQF}(D,\mathcal{P})+\lambda\mathcal{L}_{LM}^{AQF}(D,\mathcal{P})+\eta \mathcal{L}_{RQ}^{AQF}(\mathcal{P}) \label{eq:final_loss}
\end{equation}
where $\lambda$ and $\eta$ denotes the weight of the LM task and the RQ task.

\begin{table}[t]
  \centering
  \small
    \begin{tabular}{c|cc}
    \toprule
     \diagbox{\small Tasks}{\small Format}     & BCQA  & BQCA (ours) \\
    \midrule
    SQuAD & 72.3  & \textbf{81.1 } \\
    SST   & 90.9  & \textbf{92.2 } \\
    SRL   & 70.4  & \textbf{73.9 } \\
    WOZ   & 84.9  & \textbf{86.7 } \\
    E2ENLG & 48.8  & \textbf{49.2 } \\
    RNNLG (rest) & 64.0  & \textbf{64.6 } \\
    RNNLG (hotel) & 65.4  & \textbf{66.4 } \\
    RNNLG (tv) & 70.8  & \textbf{71.6 } \\
    RNNLG (laptop) & 73.0  & \textbf{73.2 } \\
    \bottomrule
    \end{tabular}%
	\caption{The results on each task with single task learning (only the QA task is trained). The format "\textit{BCQA}" was applied in previous LAMOL-based methods. Better performance in boldface. }
  \label{tab:single-task}%
\end{table}%

\section{Experiment Setup}
\subsection{Datasets}
To be comparable with previous works, we chose four different kinds of tasks from DecaNLP \citep{decaNLP} and five sequence-generation tasks \citep{l2kd} from different domains. Details are summarized in the Table \ref{tab:dataset} in Appendix \ref{sec:app-datasets}. The SQuAD is a question-answering dataset with 12 main types of questions and other countless types of questions. SST is a sentiment analysis dataset with one question. SRL is a semantic role labeling dataset with 6 types of questions. WOZ is a goal-oriented dialogue dataset with one question. E2ENLG \citep{e2enlg} and RNNLG (rest/hotel/tv/laptop) \citep{rnnlg}are sequence generation tasks with one question for different domains.

\subsection{Baselines}

(i) \textbf{Finetune}: Finetune GPT-2 individually based on the task order. 
(ii) \textbf{LAMOL} \cite{sun2019lamol}: Training the model with the "\textit{BCQA}" format data. LAMOL$_G$ and LAMOL$_T$ indicate that the begin token is the task-independent token [GEN] and the ask-specified token [TASK], respectively. LAMOL$_R$ indicates that the model replays real data from previous tasks instead of pseudo data. 
(iii) \textbf{L2KD} \cite{l2kd}: An improved version of LAMOL with distillation. Firstly, training a single-task model on the new task to obtain the teacher model. Then the teacher model is distilled to the model trained on the previous tasks. This model uses the task-specified token. All the LAMOL-based baselines were studied upon the "\textit{BCQA}" format of LAMOL. We only select L2KD on behalf of other LAMOL-based baselines because of a lack of computation resources. 
(iv) \textbf{Multitask}: Training all tasks simultaneously. Multitask learning is often regarded as the upper bound of lifelong learning.

\subsection{Implementation Detail}
For fairness comparation. we implement experiments following LAMOL\footnote{\url{https://github.com/jojotenya/LAMOL}}. GPT-2 with 12 layers is selected as the language model. All experiments are run on a single Tesla P100 (12GB). Each task is trained for 9 epochs. The pseudo data is generated by greedy deocde. The other hyperparameter settings are the same as for LAMOL and are detailed in the Table \ref{tab:hyper} in the Appendix \ref{sec:app-hyper}.

\begin{table*}[tp]
  \centering
\small
    \begin{tabular}{c|cccccccc}
    \toprule	
    
    Methods & \scriptsize SST SRL WOZ & \scriptsize SST WOZ SRL &\scriptsize SRL SST WOZ &\scriptsize SRL WOZ SST &\scriptsize WOZ SST SRL & \multicolumn{1}{c|}{\scriptsize WOZ SRL SST} & Avg. & Std. \\
    \midrule
    Finetune & 42.8  & 25.2  & 58.8  & 32.2  & 25.6  & \multicolumn{1}{c|}{36.2} & 36.8 & 11.6 \\
    \midrule
    LAMOL$_T^{0.05}$ & 77.3  & 76.9  & 78.1  & 74.7  & 73.4  & \multicolumn{1}{c|}{75.8} & 76.0 &1.6 \\
    LAMOL$_G^{0.05}$ & 79.6  & 78.9  & 73.1  & 73.7  & 68.6  & \multicolumn{1}{c|}{75.7} & 74.9  &3.4\\
    AQF-RQ$_T^{0.05}$ & 81.6  & 80.8  & \underline{80.3}  & \underline{80.0}  & \underline{80.6}  & \multicolumn{1}{c|}{80.7} & \underline{80.7} &\underline{1.1}\\
    AQF-RQ$_G^{0.05}$ & \underline{82.2}  & \underline{81.3}  & 78.9  & 79.0  & 79.2  & \multicolumn{1}{c|}{\underline{81.7} } & 80.4  &2.9\\
    \midrule
    LAMOL$_T^{0.2}$ & 79.4  & 79.9  & 80.1  & 78.7  & 79.8  & \multicolumn{1}{c|}{79.0} & 79.5 &0.5 \\
    LAMOL$_G^{0.2}$ & 80.0  & 80.7  & 79.6  & 78.7  & 78.4  & \multicolumn{1}{c|}{80.5} & 79.7 &0.8 \\
    AQF-RQ$_T^{0.2}$ & 82.2  & \textbf{83.0} & \textbf{81.4} & \textbf{81.6} & 82.2  & \multicolumn{1}{c|}{\textbf{82.9}} & \textbf{82.2} &\textbf{0.5}\\
    AQF-RQ$_G^{0.2}$ & \textbf{82.8} & 82.7  & 80.9  & 80.9  & \textbf{82.3} & \multicolumn{1}{c|}{82.4} & 82.0  &0.8\\
    \midrule
    Multitask & \multicolumn{7}{c}{82.5} \\
    \bottomrule
    \end{tabular}%
    \caption{The results on [SST, SRL, and WOZ]. Each column is the average score of a task order on three tasks. Avg. means that the average score on 6 task orders of [SST, SRL, and WOZ]. In the column "Methods", The subscripts $T$ and $G$ indicate the task-specific token [TASK] and task-independent token [GEN], respectively. The superscript indicates the value of sampling ratio $\gamma$.}
  \label{tab:deca3}%
\end{table*}%

\section{Experiments}
In this section, we experiment our proposed Ask Question First and Replay Question (AQF-RQ) an on three settings: (1) single task learning in Section \ref{sec:single-task}; (2) different types of tasks in Section \ref{sec:deca3}; (3) the same types of tasks in different domains in Section~\ref{sec:nlg}. Finally, we explore the effectiveness of our proposed AQF in Section \ref{sec:effeciveness of aqf}.

\subsection{Single Task}\label{sec:single-task}
To validate our proposed AQF that format all task into new data format "\textit{BQCA}", we experiment on each dataset independently. Each task is only trained by the QA task. As shown in Table \ref{tab:single-task}, our proposed "\textit{BQCA}" format beats "\textit{BCQA}" format which is applied in previous LAMOL-based methods on each task. It exhibits that our "\textit{BQCA}" format has the ability to increase the performance of a variety of different kinds of tasks, particularly those that were originally question-answering types (e.g. SQuAD, SRL). The SQuAD is improved significantly with the "\textit{BQCA}" format scoring 8.8 percentage points higher than the "\textit{BCQA}" format. For SRL, 3.5 percentage points of improvement come from the "\textit{BQCA}" format compared with the "\textit{BCQA}" format. For SST and WOZ, the "\textit{BQCA}" format can improve by 1.3 and 1.8 percentage points respectively. For five generation tasks, "\textit{BQCA}" format is slightly better than "\textit{BCQA}".

There are two reasons why our proposed "\textit{BQCA}" format can improve the aforementioned tasks to a certain extent: (1) Due to the characteristics of autoregression and the mask attention mechanism of the Transformers Decoder, first \textit{Q} and then \textit{C} can enable the model to obtain more accurate attention information from \textit{C} based on \textit{Q}. (2) The \textit{A} is right next to the \textit{C}, which reduces the information loss caused by the excessive length of the historical text in comparison to "\textit{BCQA}" where the \textit{A} and the \textit{C} are gapped by the \textit{Q}.

\subsection{Different Types of Tasks}\label{sec:deca3}

For a fair comparison, we conduct experiments on three different tasks in DecaNLP following LAMOL \citep{sun2019lamol}. To observe the performance in the case of sufficient and insufficient pseudo data, we select the sampling ratio $\gamma=0.2$ and $\gamma=0.05$ for experiments. Meanwhile, we conduct experiments on the condition of task-specific token [TASK] (denoted by the subscript $T$) and task-independent token [GEN] (denoted by the subscript $G$) to verify if our proposed AQF-RQ is robust when the task boundary is clear or unclear.

\begin{table*}[t]
  \centering
	\small
  
    \begin{tabular}{c|cccccc}
    \toprule
    \multirow{2}[4]{*}{Methods} & \scriptsize SST SRL WOZ & \scriptsize SST WOZ SRL &\scriptsize SRL SST WOZ &\scriptsize SRL WOZ SST &\scriptsize WOZ SST SRL & \scriptsize WOZ SRL SST \\
\cmidrule{2-7}          & \multicolumn{6}{c}{$|ps_1|:|ps_2|$} \\
    \midrule
    LAMOL$_G^{0.05}$ & 27:99 & 23:297 & 7:119 & 3:343 & 19:301 & 6:340 \\
    AQF-RQ$_G^{0.05}$ & 17:109 & 12:308 & 22:104 & 29:317 & 70:250 & 168:178 \\
    LAMOL$_G^{0.2}$ & 292:216 & 319:963 & 128:380 & 284:1100 & 225:1057 & 388:996 \\
    AQF-RQ$_G^{0.2}$ & 253:255 & 255:1027 & 239:269 & 603:781 & 712:570 & 648:736 \\
    \midrule
    \midrule
    \multirow{2}[4]{*}{Methods} & \scriptsize SST SRL WOZ & \scriptsize SST WOZ SRL &\scriptsize SRL SST WOZ &\scriptsize SRL WOZ SST &\scriptsize WOZ SST SRL & \scriptsize WOZ SRL SST \\
\cmidrule{2-7}          & \multicolumn{6}{c}{$|ps_1^{c}|:|ps_1^{nc}|/|ps_2^{nc}|:|ps_2^{c}|$} \\
    \midrule
    LAMOL$_T^{0.05}$ & 54:9/1:62 & 11:149/2:158 & 51:12/0:63 & 22:151/1:172 & 60:100/0:160 & 139:34/0:173 \\
    AQF-RQ$_T^{0.05}$ & 63:0/0:63 & 160:0/0:160 & 63:0/2:61 & 173:0/0:173 & 160:0/0:160 & 172:1/1:172 \\
    LAMOL$_T^{0.2}$ & 247:7/9:245 & 478:163/5:636 & 250:4/0:254 & 537:155/6:686 & 635:6/4:637 & 678:14/5:687 \\
    AQF-RQ$_T^{0.2}$ & 254:0/0:254 & 641:0/0:641 & 254:0/0:254 & 692:0/0:692 & 641:0/0:641 & 692:0/0:692 \\
    \bottomrule
    \end{tabular}%
\caption{The results of pseudo data distribution after learning two tasks on [SST, SRL, and WOZ]. $ps_{i}$ represents the pseudo data of $i$-th task. The subscripts $c$ and $nc$ indicate whether the pseudo data is correspond to the task or not.}
  \label{tab:deca3-ps}%
\end{table*}%

\subsubsection{Performance}\label{sec:deca3-performance}
As shown in Table \ref{tab:deca3}, Finetune, which is a baseline for other methods, suffers serious catastrophic forgetting and has large gap bewteen multitask learning. 
Let's first observe the situation where the pseudo data is sufficient (i.e. $\gamma=0.2$). LAMOL$_G^{0.2}$ and LAMOL$_T^{0.2}$ have good performance and are similar, indicating that in this case, the "\textit{BCQA}" format is robust to clear and unclear task boundaries, but still 3 percentage points lower than MTL. However, AQF-RQ does better than LAMOL by 2.7 percentage points when the boundary is clear and by 2.3 percentage points when the boundary is not clear. AQF-RQ is only 0.3-0.6\% worse than MTL. It is worth noting that AQF-RQ outperforms multi-task learning in the three task orders: SST-SRL-WOZ, SST-WOZ-SRL, and WOZ-SRL-SST. This indicates that AQF-RQ can not only better alleviate catastrophic forgetting but also further strengthen forward transfer between tasks.

When the pseudo data is insufficient (i.e. $\gamma=0.05$), the performance of LAMOL drops significantly, falling 6.5-7.6 percentage points lower than that of multitask learning. Furthermore, LAMOL$_T^{0.05}$ differs from LAMOL$_G^{0.05}$, indicating that LAMOL is not robust enough for clear and unclear task boundaries when the amount of pseudo data is insufficient. Although LAMOL's job boundary is apparent, our proposed AQF-RQ can improve by 4.7 and 5.5 percentage points compared to LAMOL. This illustrates that even when the amount of pseudo data is insufficient, AQF-RQ is still robust to both clear and unclear task boundaries, as evidenced by AQF-RQ$_T^{0.05}$ and AQF-RQ$_G^{0.05}$. It is also worth noting that AQF-RQ$_T^{0.05}>$ AQF-RQ$_G^{0.05}>$LAMOL$_G^{0.2}>$LAMOL$_T^{0.2}$, which means that AQF-RQ is more data-efficient and still performs better than LAMOL with a 75\% reduction in the amount of pseudo data.

\begin{table*}[t]
  \centering
\small
    \begin{tabular}{c|ccccccc}
    \toprule
    Methods & \scriptsize SST SRL WOZ & \scriptsize SST WOZ SRL & \scriptsize SRL SST WOZ & \scriptsize SRL WOZ SST & \scriptsize WOZ SST SRL & \multicolumn{1}{c|}{\scriptsize WOZ SRL SST} & Avg. \\
    \midrule
    LAMOL$_T$ & 79.4  & 79.9  & 80.1  & 78.7  & 79.8  & \multicolumn{1}{c|}{79.0} & 79.5  \\
    \multicolumn{1}{r|}{w/ AQF} & \textbf{82.1} & \textbf{83.0 } & \textbf{81.3} & \textbf{80.9} & \textbf{80.7} & \multicolumn{1}{c|}{\textbf{81.2}} & \textbf{81.5} \\
    \midrule
    LAMOL$_G$ & 80.0  & 80.7  & 79.6  & 78.7  & 78.4  & \multicolumn{1}{c|}{80.5} & 79.7  \\
    \multicolumn{1}{r|}{w/ AQF} & \textbf{82.3} & \textbf{81.5} & \textbf{81.3} & \textbf{80.4} & \textbf{78.8} & \multicolumn{1}{c|}{\textbf{81.3}} & \textbf{80.9} \\
    \midrule
    LAMOL$_R$ & 81.8  & 80.6  & 81.6  & 81.2  & 80.4  & \multicolumn{1}{c|}{80.5 } & 81.0  \\
    \multicolumn{1}{r|}{w/ AQF} & \textbf{82.5} & \textbf{81.9} & \textbf{82.3} & \textbf{82.4} & \textbf{82.7} & \multicolumn{1}{c|}{\textbf{82.7}} & \textbf{82.4} \\
    \midrule
    L2KD  & 80.1  & 79.6  & 79.5  & 79.7  & 79.9  & \multicolumn{1}{c|}{\textbf{80.4 }} & 79.9  \\
    \multicolumn{1}{r|}{w/ AQF} & \textbf{81.9} & \textbf{82.4} & \textbf{80.3} & \textbf{80.8} & \textbf{80.1} & \multicolumn{1}{c|}{80.1} & \textbf{80.9} \\
    \midrule
    \multicolumn{1}{c|}{Multitask$^*$} & \multicolumn{7}{c}{81.5} \\
    \multicolumn{1}{r|}{w/ AQF} & \multicolumn{7}{c}{\textbf{82.5}} \\
    \bottomrule
    \end{tabular}%
	\caption{The results of effectiveness of the Ask Question First (AQF) on [SST, SRL, and WOZ] when sampling ratio $\gamma=0.2$. Multitask$^*$ represents that model applies the "\textit{BCQA}" format for multitask learning.}
  \label{tab:deac3+aqf}%
\end{table*}

\subsubsection{Distribution of Pseudo Data}
The another main objective of our proposed AQF-RQ is to generate more corresponding-task pseudo data. Ideally, the ratio of pseudo data for each task is 1:1, which means $|ps_i|=|ps_j|,i \neq j$. We can determine which task the pseudo data belongs to according to the \textit{Q} because the \textit{Q} for various tasks is distinct (see Table \ref{tab:dataset}). For task-specific token [TASK], a pseudo sample is task-corresponding if the \textit{Q} is correspond to the [TASK]. For task-indepedent token [GEN], we can not judge whether a pseudo sample corresponds to the task because of unclear task boundary. In this paper, we applied this \textit{Q}-based judgment method to count the distribution of pseudo data after learning the second task. 

As shown in Table \ref{tab:deca3-ps}, AQF-RQ can generate more pseudo data of the first task for most task orders when applying task-indepedent token [GEN]. Looking at the Table \ref{tab:deca3} together, it can be observed that the performance will be better if the ratio of the pseudo data of different tasks is closer to 1:1. For SRL-SST-WOZ, SRL-WOZ-SST, WOZ-SST-SRL, and WOZ-SRL-SST, AQF-RQ generates more pseudo data than LAMOL for the first task; hence, AQF-RQ outperforms LAMOL significantly. When $\gamma=0.05$, AQF-RQ is 5.3-10.6 percentage points higher than LAMOL. When $\gamma=0.05$, AQF-RQ is 1.3-3.9 percentage points higher than LAMOL. This is due to the fact that, under the AQF-proposed "\textit{BQCA}" format, \textit{Q} can be used to help the model pay more attention to the useful details in \textit{C}, and then \textit{A} being immediately next to \textit{C} can further generate \textit{A} more accurately.

The benefit of AQF-RQ to produce pseudo data is also significant when the sentence starts with the task-specific token [TASK]. As shown in the Table \ref{tab:deca3-ps}, the pseudo samples generated by AQF-RQ are task-corresponding in the majority of task orders, except SRL-SST-WOZ and WOZ-SRL-SST, where one or two pseudo samples are not task-corresponding. The amount of not task-corresponding pseudo samples in AQF-RQ is substantially smaller than in LAMOL, giving AQF-RQ a 1.3-3.9 percentage point advantage over LAMOL.

The above experimental results demonstrate that not only does AQF-RQ generate better pseudo-data, but the "\textit{BQCA}" format is also more data-efficient and robust for lifelong language learning.

\subsection{Generation for Different Domains}\label{sec:nlg}
We conducted experiments on five sequence generation tasks for different domains by comparing LAMOL$_T$, LAMOL$_G$, and L2KD with AQF-RQ. The result is similar to Section \ref{sec:deca3-performance} and is detailed in Appendix \ref{sec:app-nlg}. AQF-RQ is not only better than LAMOL-based baselines on performance but can also forward/backward transfer more knowledge among the same task but different domains.


\subsection{Effectiveness of the Ask Question First}\label{sec:effeciveness of aqf}
To verify the effectiveness of our proposed AQF, we experimentally apply AQF to each baseline: LAMOL$_T$, LAMOL$_G$, LAMOL$_R$, and L2KD. In Section~\ref{sec:aqf}, we stated that AQF is used better when the pseudo data is sufficient. Therefore, we set $\gamma=0.2$ for experiments. This value is also the best setting for those baselines.

As shown in the Table \ref{tab:deac3+aqf}, applying AQF to each baseline resulted in varying degrees of improvement. LAMOL$_T$ has the most noticeable improvement, with a 2 percentage point increase. This demonstrates that having the \textit{A} close to the \textit{C} using the "\textit{BQCA}" format proposed by AQF can reduce information attenuation caused by the \textit{A} being far away from the \textit{C}. For example, the \textit{C} and the \textit{A} are separated by the \textit{Q} in the "\textit{BCQA}" format. This benefit is more obvious when the task boundary is clear (i.e. the sentence starts with [TASK]). [TASK] can tighten the bond between "\textit{BQC}" and "\textit{A}". 

Applying AQF improves LAMOL$_G$ by 1.2 percentage points, which is less than LAMOL $_T$. It is because the generated pseudo data is still biased toward new tasks due to the unclear task boundary (the sentence begins with the task-independent token [GEN]). However, thanks to the \textit{A} being next to the \textit{C} in the "\textit{BQCA}" format, it is still capable of learning from a small amount of pseudo data. 

For LAMOL$_R$, AQF can enhance the model by 1.4 percentage points, which is only 0.1\% behind multi-task learning. LAMOL$_R$ uses the real data of the old task instead of the generated pseudo data when learning new tasks, so this further demonstrates that the "\textit{BQCA}"format proposed by AQF is more conducive to lifelong language learning: it can use \textit{Q} to help the model pay more attention to the important information in \textit{C}, then further makes the generation of \textit{A} more accurate. 

For L2KD, the application of AQF improves the model by one percentage point, demonstrating that AQF is still applicable to the previous LAMOL-based enhanced approaches and is capable of effectively applied to other LAMOL-based studies.

For MTL, AQF can bring an improvement of 1 percentage point. At the same time, the improvement of each baseline demonstrates that the improvement brought by AQF is comprehensive: it is not only conducive to alleviating catastrophic forgetting but also improves the upper bound of the model. Therefore, we believe that AQF has the potential to apply LLL to real-world scenarios.

\section{Effectiveness of the Replay Questions}
Since the Replay QuestionsRQ is proposed based on AQF, we verify and analyze the effectiveness of the RQ based on Table \ref{tab:deca3} and \ref{tab:deac3+aqf}. The AQF-RQ performance in Table \ref{tab:deca3} subtracts the corresponding LAMOL+AQF performance in Table \ref{tab:deac3+aqf} is the gains obtained by AQF-RQ come from the RQ step. Therefore, RQ$_{G}^{0.2}$ = AQF-RQ$_{G}^{0.2}$ - LAMOL+AQF$_{G}^{0.2}$ = 1.1, and RQ$_{T}^{0.2}$ = AQF-RQ$_{T}^{0.2}$ - LAMOL+AQF$_{T}^{0.2}$ = 0.7. It demonstrates that the RQ step based on the AQF can help the model generate more task-corresponding pseudo samples by making $P(Q_i|B_i)>P(Q_{j \neq i}|B_i)$.

\section{Conclusion and Future Work}
This work proposed AQF-RQ which is a simple yet efficient and robust lifelong language learning method. We propose a new question-first data format that is train-efficient without additional computational resources and new parameters. In AQF's "\textit{BQCA}" format, generating \textit{Q} first makes it easier to generate task-corresponding pseudo-data, and \textit{A} is more accurate because \textit{A} is next to \textit{C}. In addition, RQ can strengthen the model's attention to the problem, so that the model has a greater probability of generating task-corresponding questions, and further generates more task-corresponding pseudo-data. AQF-RQ effectively alleviates catastrophic forgetting, only 0.36\% lower than multi-task learning. Due to a lack of computing resources, we did not experiment on larger datasets and longer task orders, which we leave as future work. In addition, we will investigate ways to enhance performance when the task boundary is unclear.

\section*{Acknowledgements}
This work has been supported by The Youth Innovation Promotion Association of the Chinese Academy of Sciences
(E1291902), Jun Zhou (2021025).

\bibliography{anthology,custom}
\bibliographystyle{acl_natbib}

\appendix

\section{Datasets}\label{sec:app-datasets}
As we can not obtain test set of SQuAD since it is hidden from the host, we use development set for testing. For other tasks, we use the corresponding test set.
A normalized F1 (nF1) metric that lower text and remove punctuation and articles, is used to evaluate SQuAD and SRL. The exact match (EM) is used to evaluate SST. The turn-based dialogue state EM (dsEM) is used for WOZ. The ROUGE is used to evaluate E2ENLG and RNNLG (rest/hotel/tv/laptop)The size of each dataset is detailed in Table \ref{tab:dataset}.

As shown in Table \ref{tab:dataset}, SST, WOZ, E2ENLG and RNNLG (rest/hotel/tv/laptop) only have one question. The SQuAD has mainly 12 types of questions and other countless types of questions. SRL has 6 types of questions.

\section{Hyperparameter}\label{sec:app-hyper}
The main hyperparameters in our experiments are detailed in Table \ref{tab:hyper}.

\begin{table}[t]
\centering
\begin{tabular}{@{}ll@{}}
\toprule
\textbf{hyperparameter} & \textbf{value} \\ \midrule
optimizer               & AdamW          \\
adam epsilon            & $1\times10^{-4}$           \\
learning rate           & $1\times10^{-4}$           \\
weight of RQ task           & $\eta = 0.2$            \\
weight of LM task           & $\lambda =0.25$           \\
max gradient norm       & 1.0            \\
learning rate schedule  & warmup linear  \\
warmup ratio            & 0.005          \\
max length              & 1024            \\
top-k sampling          & k=20           \\ \bottomrule
\end{tabular}
\caption{The main hyperparameters in our experiment.}
\label{tab:hyper}%
\end{table}

\begin{table*}[htbp]
  \centering
  
    \begin{tabular}{lcccc|l}
    \toprule
    \textbf{Dataset} & \textbf{\#Train} & \textbf{\#Test} & \textbf{Metric} & \textbf{Question Type} & \multicolumn{1}{c}{\textbf{Question}} \\
    \midrule
    \multicolumn{5}{l|}{\textit{different types of tasks}} &  \\
    SQuAD & 87599 & 10570 & nF1   & 12+$\infty$ &  \makecell[l]{what/who/whose/whom/when/ \\where/how/why/which/if/do/is … \\+ other countless types of questions} \\
    \midrule
    SST   & 6920  & 1821  & EM    & 1     & is this review negative or positive?  \\
    \midrule
    SRL   & 6414  & 2201  & nF1   & 6     & \makecell[l]{what/who/whose/whom/ \\when/where/how/why …} \\
    \midrule
    WOZ   & 2536  & 1646  & dsEM  & 1     & what is the change in state? \\
    \midrule
    \multicolumn{5}{l|}{\textit{suquence generation for different domains}} &  \\
    E2ENLG & 6000  & 2000  & \multirow{5}[1]{*}{ROUGE} & \multirow{5}[1]{*}{1} & \multirow{5}[1]{*}{what is the natural language form?} \\
    RNNLG (rest) & 6228  & 1039  &       &       &  \\
    RNNLG (hotel) & 6446  & 1075  &       &       &  \\
    RNNLG (tv) & 8442  & 1407  &       &       &  \\
    RNNLG (laptop) & 7944  & 2649  &       &       &  \\
    \bottomrule
    \end{tabular}%
\caption{The summarized results for datasets. }
  \label{tab:dataset}%
\end{table*}%

\section{Sequence Generation for Different Domains}\label{sec:app-nlg}

\begin{figure*}[ht]
\centering 
\includegraphics[scale=0.7]{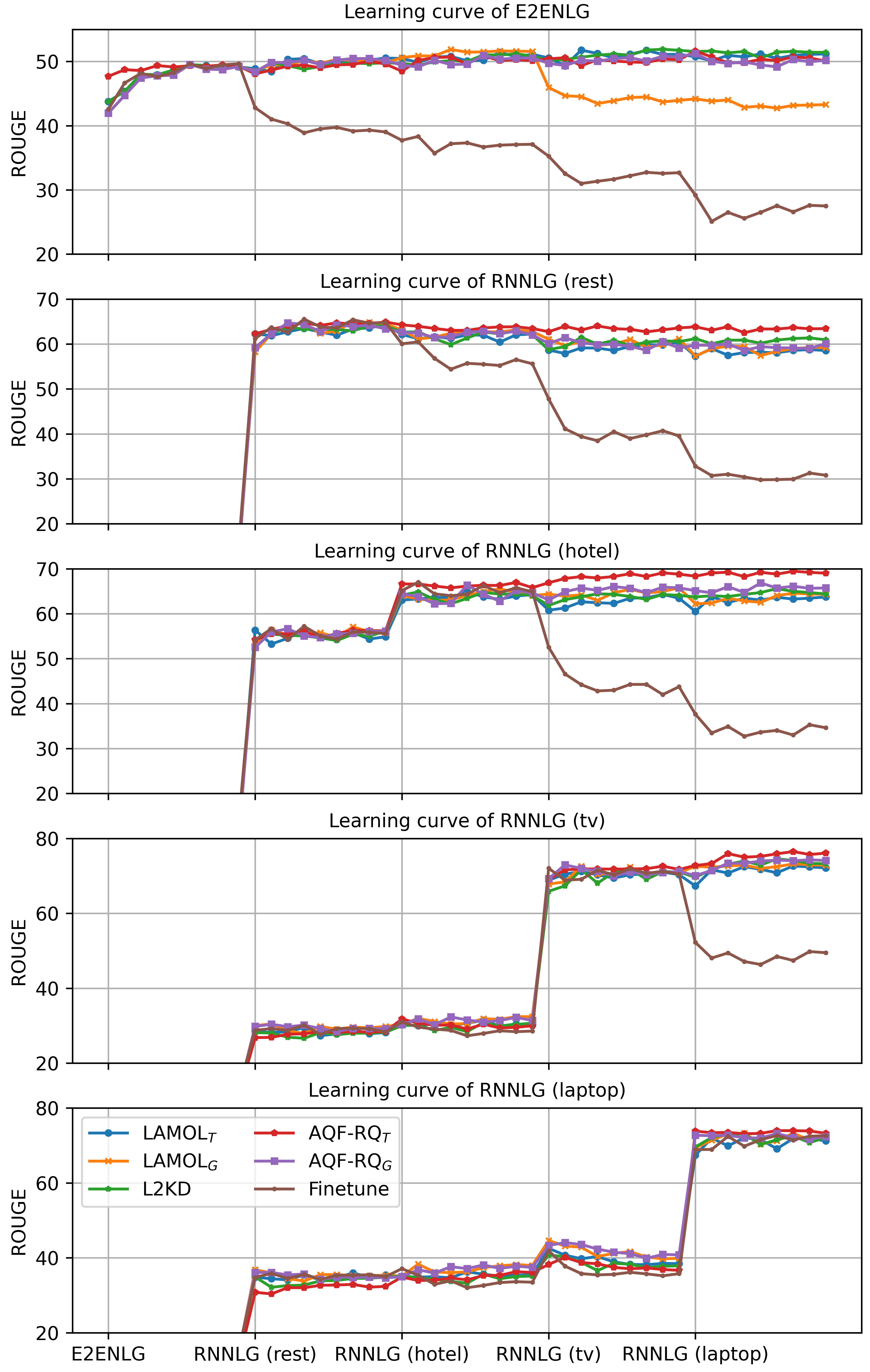} 
\caption{Results on sequence generation tasks for five different domains when $\gamma=0.2$.}
\label{fig:nlg}
\end{figure*}

We conducted experiments on five sequence generation tasks for different domains by comparing LAMOL$_T$, LAMOL$_G$, and L2KD with AQF-RQ. Experiments are conducted with sufficient pseudo data (i.e. $\gamma$=0.2). Following L2KD, we select the task order from hard to easy: E2ENLG-RNNLG (rest) - RNNLG (hotel) - RNNLG (tv) - RNNLG (laptop). As shown in the Fig.\ref{fig:nlg}, we can observe that all models perform well except for Finetune and LAMOL$_G$. However, AQF-RQ can still outperform other baselines. When learning the same type of tasks in different domains, first we expect that previous tasks can promote the learning of new tasks. This is known as the forward transfer. In terms of the forward transfer, AQF-RQ is better at acquiring knowledge from previous tasks that is beneficial to new tasks. The second is backward transfer, which refers to learning new tasks while consolidating and strengthening previous tasks. It can be observed that AQF-RQ also has better backward transfer. For example, from the fourth task RNNLG.tv on, the backward transfer of AQF-RQ on the previous tasks is higher than other baselines. 
As the number of learned tasks increases, the unclear task boundary gradually becomes significantly weaker than the clear task boundary. For LAMOL$_G$, the performance on E2ENLG begins to drop significantly from the fourth task RNNLG.tv. But our AQF-RQ$_G$ remains on an upward trend. According to the above analysis, it can be concluded that AQF-RQ has stronger forward transfer and backward transfer capabilities in different domains of the same type of task.

\end{document}